%% file: iclr2023_conference.tex
\documentclass{article} % For LaTeX2e
\usepackage{iclr2023_conference,times}

\input{math_commands.tex}

\usepackage{hyperref}
\usepackage{url}
\usepackage{graphicx}
\usepackage{algorithm}
\usepackage{algpseudocode}
\usepackage{amsmath}
\usepackage{bm}
\algnewcommand\algorithmicforeach{\textbf{for each}}
\algdef{S}[FOR]{ForEach}[1]{\algorithmicforeach\ #1\ \algorithmicdo}

\title{Improve State-Level Wheat Yield Forecasts in Kazakhstan on GEOGLAM's EO Data by \\Leveraging A Simple Spatial-Aware Technique
}

\author{Anh Nhat Nhu \\
Department of Computer Science \\
University of Maryland \\
NASA Harvest \\
College Park, MD 20742, USA \\
\texttt{anhu@terpmail.umd.edu} \\
\And
Ritvik Sahajpal \\
Department of Geographical Science \\
University of Maryland \\
NASA Harvest \\
College Park, MD 20742, USA \\
\texttt{ritvik@umd.edu} \\
\And
Christina Justice \\
Department of Geographical Science \\
University of Maryland \\
NASA Harvest \\
College Park, MD 20742, USA \\
\texttt{justice@umd.edu} \\
\And
Inbal Becker-Reshef \\
Department of Geographical Science \\
University of Maryland \\
NASA Harvest \\
College Park, MD 20742, USA \\
\texttt{ireshef@umd.edu} \\
}

\iclrfinalcopy % Uncomment for camera-ready version, but NOT for submission.
\begin{document}

\maketitle

\begin{abstract}
 Accurate yield forecasting is essential for making informed policies and long-term decisions for food security. Earth Observation (EO) data and machine learning algorithms play a key role in providing a comprehensive and timely view of crop conditions from field to national scales. However, machine learning algorithms' prediction accuracy is often harmed by spatial heterogeneity caused by exogenous factors not reflected in remote sensing data, such as differences in crop management strategies. In this paper, we propose and investigate a simple technique called state-wise additive bias to explicitly address the cross-region yield heterogeneity in Kazakhstan. Compared to baseline machine learning models (Random Forest, CatBoost, XGBoost), our method reduces the overall RMSE by 8.9\% and the highest state-wise RMSE by 28.37\%. The effectiveness of state-wise additive bias indicates machine learning's performance can be significantly improved by explicitly addressing the spatial heterogeneity, motivating future work on spatial-aware machine learning algorithms for yield forecasts as well as for general geospatial forecasting problems.
\end{abstract}

%%%%%%%%%%%%%%%%%%%%%%%%%%%%%%%%%%%%%%%%%%%%%%%%%%%%%%%%%%%%%%%%%%
%%%%%%%%%% SECTION 1: INTRODUCTION %%%%%%%%%%%%%%%%%%%%%%%%%%%%%%%
%%%%%%%%%%%%%%%%%%%%%%%%%%%%%%%%%%%%%%%%%%%%%%%%%%%%%%%%%%%%%%%%%%
\section{Introduction}
Accurate crop yield forecasts can benefit governments, policymakers, and individual farmers by providing better insights into various exogenous drivers that impact the 
agricultural markets. These insights can lead to earlier responses and better-informed decisions to improve food security at both regional and international scales \citep{becker2022nasa}. Recently, machine learning algorithms have been applied on Earth Observation (EO) data and have shown a great potential to improve the reliability of these forecasts \citep{basso2019seasonal}.

In this paper, we consider the use of EO data collected from the GEOGLAM Crop Monitor AgMet System (https://cropmonitor.org) and tree-based algorithms to directly forecast wheat yields in Kazakhstan, the $10^{th}$ largest wheat exporter in the world \citep{FAOSTAT2022}. A prominent challenge negatively impacting Machine Learning models' performance in forecasting yields is the spatial yield heterogeneity due to exogenous factors like local farming practices or crop varietals that are not reflected in remote sensing data. \cite{lee2022maize} proposed to train a separate model for each province, successfully reducing the state-wise prediction errors. However, in our dataset, due to a very small amount of yield data available for each province (typically less than 20 data points), this approach results in highly unreliable and overfit models with error rates far exceeding those of baseline models, as shown in Figure~\ref{fig:boxplot}. To improve upon this issue, we focus on reducing the errors, especially in provinces with the least accurate yield predictions, by using state-wise additive bias. First, we followed the methodologies in \cite{sahajpal-argentina} to create features from EO data and investigate the performance of various baseline tree-based models, including XGBoost, CatBoost, and Random Forest, in forecasting wheat yields at the state level. Next, each state-wise additive bias was separately added to the model's predictions in each province to obtain the final yield forecast. This approach shows a remarkable increase in overall performance, with the most significant benefits being seen in the province with the highest baseline yield errors (Almatinskaya). Furthermore, since state-wise bias adds no computational overhead during the inference process, this technique can be efficiently applied to improve yield predictions in other datasets. 
% Although there are certainly many technical aspects in this paper that we could focus on to further improve the forecasts, our work serves as a groundwork to explicitly address the spatial heterogeneity and increase the reliability of machine learning models in forecasting crop yields.

%%%%%%%%%%%%%%%%%%%%%%%%%%%%%%%%%%%%%%%%%%%%%%%%%%%%%%%%%%%%%%%%%%
%%%%%%%%%% SECTION 2: DATA & METHODS %%%%%%%%%%%%%%%%%%%%%%%%%%%%%
%%%%%%%%%%%%%%%%%%%%%%%%%%%%%%%%%%%%%%%%%%%%%%%%%%%%%%%%%%%%%%%%%%
\section{Data and Methods}

\subsection{Collecting and Extracting EO Data}
We use multiple EO predictors (https://cropmonitor.org/tools/agmet/) including crop phenological information derived from the MODIS NDVI that provides a proxy for crop vigor and phenology, MODIS Leaf Area Index (LAI), temperature, precipitation, SMAP soil moisture, and evaporative stress index (ESI). These inputs are subsets to cropped areas using a wheat crop mask for Kazakhstan. The EO products used here are complementary and capture different facets of crop response to abiotic factors (temperature, precipitation, solar radiation) and its variation by phenological growth stage and geography.

\subsection{Data Preprocessing}
The EO dataset has daily data spanning from 2001 to 2020. We subset this data to the crop growth season (May - September). We use EO data (NDVI, growing degree days, daily minimum and maximum temperature, soil moisture, evaporative stress index, and precipitation) to as input features for training and evaluating machine learning models and to compute state-wise bias. We also include information on the previous season's yield and the average yield from the last 5 years as additional variables in the model. Overall, we have 75 samples for each province (15 years x 5 months in the growing season).
% Since the dataset only spans from 2001 to 2020, information of 5-year average yield are only available starting from 2006. Therefore, the final dataset only includes data from 2006 to 2020. To reduce data redundancy, we reformat our dataset from daily to monthly data by averaging the features, grouped by province, Month and Year. In other words, we only produce a single yield forecast for each region in each month in a particular year instead of redundantly forecasting yield for every day in the dataset. 

\subsection{Model Training and Evaluation}
\begin{figure}[ht]
\begin{center}
\includegraphics[width=1.0\textwidth]{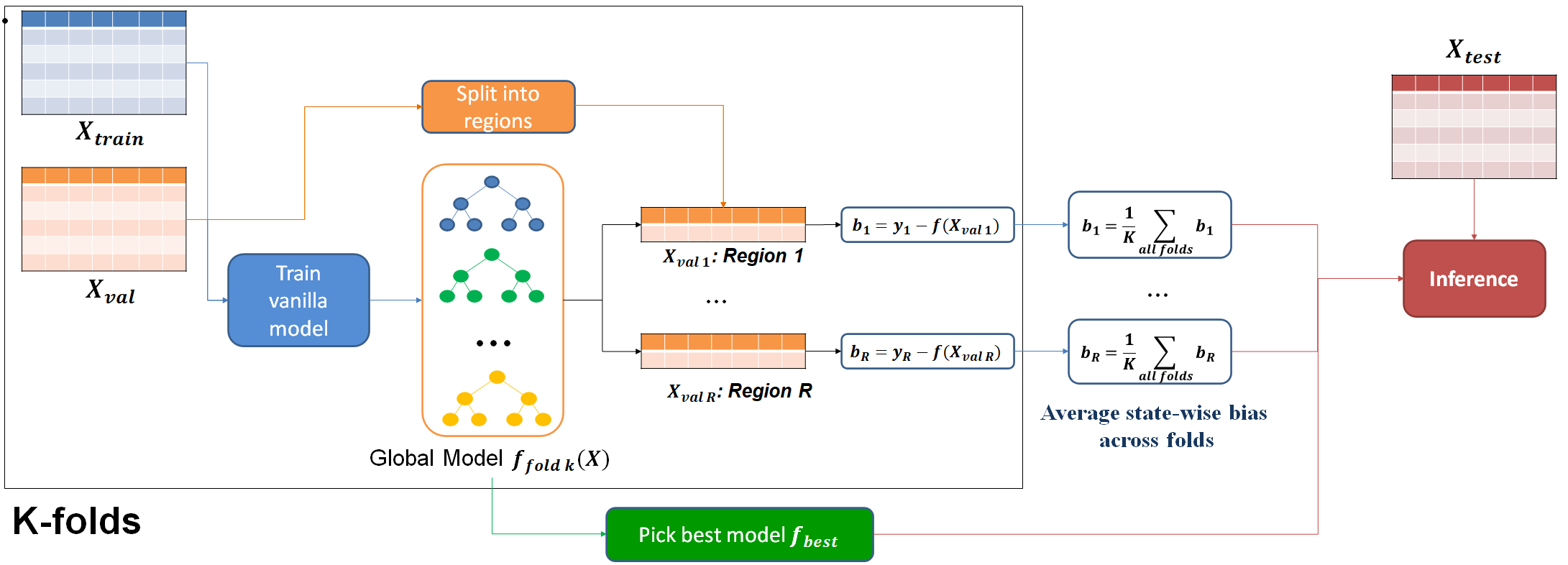}
\end{center}
\caption{Algorithmic flow of the training and evaluation process.
}
\label{fig:flow}
\end{figure}

We trained and evaluated the effectiveness of the state-wise bias by applying this bias to the baseline tree models (XGBoost, CatBoost, and Random Forest) to forecast wheat yields at the state level in Kazakhstan. The state-wise bias is automatically calculated during the training process of each model. Algorithm~\ref{alg:training} and Figure~\ref{fig:flow} present the complete training pipeline to train models and compute state-wise bias. We leave one year for testing, as suggested by \cite{MERONI2021108555}, and split the remaining data into training (10 years) and validation sets (4 years) for model optimization. To maximize the amount of data used in the training process and increase the robustness of state-wise bias to unseen data, we employed the k-fold cross-validation method each test year \citep{gmd-15-3519-2022}. In each fold, the error of each state was sampled using the corresponding validation set of the fold. The final state-wise bias of each state is the average of the recorded validation errors in all $k$ folds.

%%%%%%%%%%%%%%%%%%%%%%%%%%%%%
%%%%%% Algorithm Flow %%%%%%%
%%%%%%%%%%%%%%%%%%%%%%%%%%%%%

\begin{algorithm}
\caption{Model training and state-wise bias computation}\label{alg:training}
\hspace*{\algorithmicindent} \textbf{Input} Input features $\bm{X}$, targets $\bm{y}$ \\
\hspace*{\algorithmicindent} \textbf{Output} model $f$,  state-wise bias $b$
\begin{algorithmic}[1]
\State $(\bm{X}_{train/val}, \bm{y}_{train/val}), (\bm{X}_{test}, \bm{y}_{test}) =$ split $\bm{X}, \bm{y}$
\State
% \State SET $\bm{bias_{state}} \gets$ empty dictionary
\ForEach {$(\bm{X}_{train}, \bm{y}_{train}), (\bm{X}_{val}, \bm{y}_{val}) \in $ k-fold split}
\State Initialize model $f$
\State Fit $f(\bm{X}_{train})$,  $\bm{y}_{train}$
\State Evaluate on $f(\bm{X}_{val})$,  $\bm{y}_{val}$
\State Done training model $f$
\Comment{End training model for current fold}
\\
\State $\hat{y}_{val} = f(\bm{X}_{val})$
\ForEach{state} 
    \State state\_bias = mean$(\bm{y}_{val}\:[state] - \bm{\hat{y}}_{val}\:[state])$ \Comment{state-wise bias for current fold}
    \State $b$[state].append(state\_bias)
\EndFor 
\EndFor \Comment{End training $k$ models on $k$ folds}
\\
\ForEach{state} \Comment{Final state-wise bias for each state}
    \State $b$[state] =  mean($b$[state])
\EndFor 
\State
\State \textbf{Return} model $f$, state-wise bias $b$
\end{algorithmic}
\end{algorithm}

%%%%%%%%%%%%%%%%%%%%%%%%%%%%%
%%%%%% End Algorithm %%%%%%%
%%%%%%%%%%%%%%%%%%%%%%%%%%%%%

The fundamental motivation for computing state-wise bias is that we observed baseline models are often biased toward values close to the mean yields, underestimating high yields in provinces with high productions, as discussed in Section~\ref{sec:performance}. These high yields can be caused by factors typically not covered in satellite data, such as political and economical forces that allow some provinces to be the main wheat producer of the country. Although we have incorporated the regional information as categorical data in baseline models, the models still suffer from this bias. Therefore, state-wise bias is proposed as a simple yet effective technique to alleviate this spatial heterogeneity problem, resulting in a significant decrease in both MAPE and RMSE, as shown in Section~\ref{sec:results}
%Although we only tested and demonstrated the performance of state-wise bias with only 3 models, this technique is generally applicable to any other models for crop yield predictions. %

% \label{gen_inst}

% The text must be confined within a rectangle 5.5~inches (33~picas) wide and
% 9~inches (54~picas) long. The left margin is 1.5~inch (9~picas).
% Use 10~point type with a vertical spacing of 11~points. Times New Roman is the
% preferred typeface throughout. Paragraphs are separated by 1/2~line space,
% with no indentation.

% Paper title is 17~point, in small caps and left-aligned.
% All pages should start at 1~inch (6~picas) from the top of the page.

% Authors' names are
% set in boldface, and each name is placed above its corresponding
% address. The lead author's name is to be listed first, and
% the co-authors' names are set to follow. Authors sharing the
% same address can be on the same line.

% Please pay special attention to the instructions in section \ref{others}
% regarding figures, tables, acknowledgments, and references.

% There will be a strict upper limit of 9 pages for the main text of the initial submission, with unlimited additional pages for citations. 

%%%%%%%%%%%%%%%%%%%%%%%%%%%%%%%%%%%%%%%%%%%%%%%%%%%%%%%%%%%%%%%%%%
%%%%%%%%%% SECTION 3: RESULTS & ANALYSIS %%%%%%%%%%%%%%%%%%%%%%%%%
%%%%%%%%%%%%%%%%%%%%%%%%%%%%%%%%%%%%%%%%%%%%%%%%%%%%%%%%%%%%%%%%%%
\section{Results and Analysis}\label{sec:results}
\subsection{Model Performance}\label{sec:performance}

%%%%%%%%%%%%%%%%%%%%%%%%%%%%%
%%%% Figure: ScatterPlot %%%%
%%%%%%%%%%%%%%%%%%%%%%%%%%%%%
\begin{figure}[ht]
\begin{center}
\includegraphics[width=0.8\textwidth, height=0.5\textwidth]{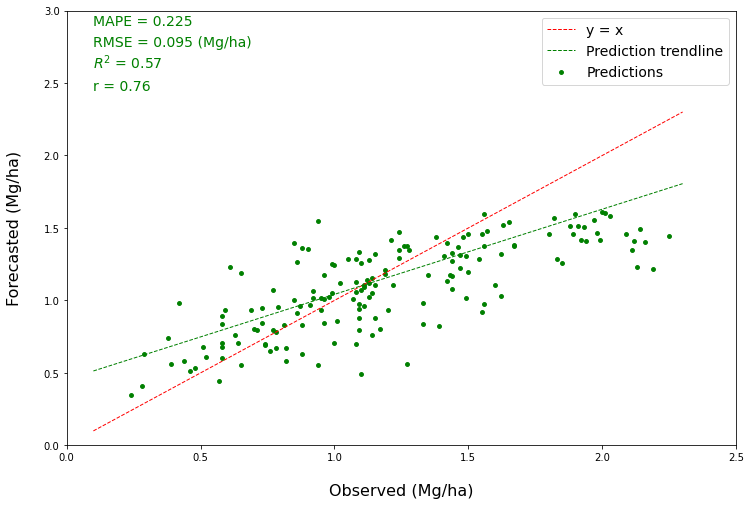}
\end{center}
\caption{Scatter plot showing relationships between multi-year predicted and ground-truth yields of different provinces using leave-one-out year testing strategy. 
%For each test year, we use different model parameters trained on a different training set excluding such test year. The RMSE, MAPE, $R^2$, and Pearson's correlation of multi-year predictions are computed using all predictions in all provinces (nationally). %
}
\label{fig:scatterplot}
\end{figure}

% We comprehensively evaluate the performance of our state-wise bias approach by separately testing on each region and in different leave-one-out test years from 2006 to 2020. The testing period is 3 months prior to the harvesting month. Since our training process used a leave-one-out year as the test year, to collect the test results on different years, we iteratively set each year as the test year and trained models on the remaining years in each iteration. Each point in the scatter plot represents the forecasted yield of a province in a particular year, and the best model (XGBoost + state-wise bias) was selected to test each corresponding test year. 
Overall, the MAPE and RMSE of XGBoost complemented with state-wise bias are \textbf{22.5\%} and \textbf{0.095 Mg/ha}, respectively. Our model explains 57\% of the yield variation in our dataset. Based on the scatter plot, we observed that the model performs well when the yields are average or low, but it consistently underestimates the yield by a large margin when yields are much higher ($\geq 1.75$ Mg/ha). Those high yields are often from provinces with the highest wheat production, such as Almatinskaya, or in exceptionally good years. 
% In future research, we will focus on conducting in-depth investigation of the root cause of such yield underestimation and developing new methods to address this problem.

%%%%%%%%%%%%%%%%%%%%%%%%%%%%%
%%%%%%%% End Figure %%%%%%%%%
%%%%%%%%%%%%%%%%%%%%%%%%%%%%%

\subsubsection{Comparison to baseline models}
To investigate the effect of state-wise bias, we test various 
 models on different out-of-fold test years and compare the performance with and without state-wise bias. Our comparison involves both full dataset evaluation (national level) and evaluation by each province (regional level). Table~\ref{table-comparison} shows that the RMSE at the national level is decreased by \textbf{8.1\%} to \textbf{9.76\%}, resulting in an overall improvement over baseline models. The most significant improvements are observed in Almatinskaya (\textbf{24.04\%} to \textbf{28.37\%}) and Yujno-Kazachstanskaya (\textbf{6.95\%} to \textbf{8.84\%}) provinces, two provinces with the highest multi-year wheat yields and highest forecasting errors. Specifically, the average multi-year wheat yields of Almatinskaya and Yujno-Kazachstanskaya are $1.793$ and $1.601$ Mg/ha, respectively, while the national average yield is only $1.103$ Mg/ha.

%%%%%%%%%%%%%%%%%%%%%%%%%%%%%
%%%%% Table: Baseline %%%%%%%
%%%%%%%%%%%%%%%%%%%%%%%%%%%%%
\begin{table}[ht]
\caption{Percentage change in RMSE of state-wise bias compared to the vanilla model (negative values represent improvements). The RMSE of each state is computed using all cross-year predictions for that state. We computed the RMSE's percentage change by subtracting the RMSE of vanilla models from the RMSE of state-wise bias, then dividing the result by the RMSE of vanilla models.
}

%%%%%%%%%%%%%%%%%%%%%%%%%%%%%
%%%%%% End: Baseline %%%%%%%%
%%%%%%%%%%%%%%%%%%%%%%%%%%%%%

\label{table-comparison}
\begin{center}
\begin{tabular}{lccc}
\multicolumn{1}{l}{\bf Province}  &\multicolumn{1}{c}{\bf XGBoost} &\multicolumn{1}{c}{\bf CatBoost} &\multicolumn{1}{c}{\bf Random Forest}
\\ \hline 
Akmolinskaya                &\bf{-0.47\%}       &+1.13\%        &\bf{-1.74\%}   \\
Aktyubinskaya               &+1.54\%            &+1.35\%        &\bf{-5.60\%}   \\
Almatinskaya                &\bf{-28.37\%}      &\bf{-24.26\%}  &\bf{-24.04\%}  \\
Jambylslkaya                &+1.35\%            &+2.49\%        &+0.08\%        \\
Karagandinskaya             &\bf{-1.37\%}       &+0.55\%        &\bf{-0.72\%}   \\
Kustanayskaya               &\bf{-3.85\%}       &\bf{-2.64\%}   &\bf{-3.55\%}   \\
Pavlodarskaya               &\bf{-0.65\%}       &+2.86\%        &\bf{-2.18\%}   \\
Severo-Kazachstanskaya      &+2.38\%            &+26.86\%       &\bf{-15.42\%}  \\
Vostochno-Kazachstanskaya   &\bf{-4.48\%}       &\bf{-4.11\%}   &\bf{-1.17\%}   \\
Yujno-Kazachstanskaya       &\bf{-8.84\%}       &\bf{-6.95\%}   &\bf{-7.69\%}   \\
Zapadno-Kazachstanskaya     &\bf{-1.38\%}       &+0.14\%        &\bf{-0.62\%}   \\
\hline
National                    &\bf{-8.90\%}       &\bf{-8.10\%}   &\bf{-9.76\%}   \\

%%%%%%%%%%%%%%%%%%%%%%%%%%%%%
%%%%%%%% End Table %%%%%%%%%%
%%%%%%%%%%%%%%%%%%%%%%%%%%%%%

\end{tabular}
\end{center}
\end{table}

\subsubsection{Comparison to region-specific models}

%%%%%%%%%%%%%%%%%%%%%%%%%%%%%
%%%%% Figure: Boxplots %%%%%%
%%%%%%%%%%%%%%%%%%%%%%%%%%%%%
\begin{figure}[ht]
\begin{center}
\includegraphics[width=0.6\textwidth]{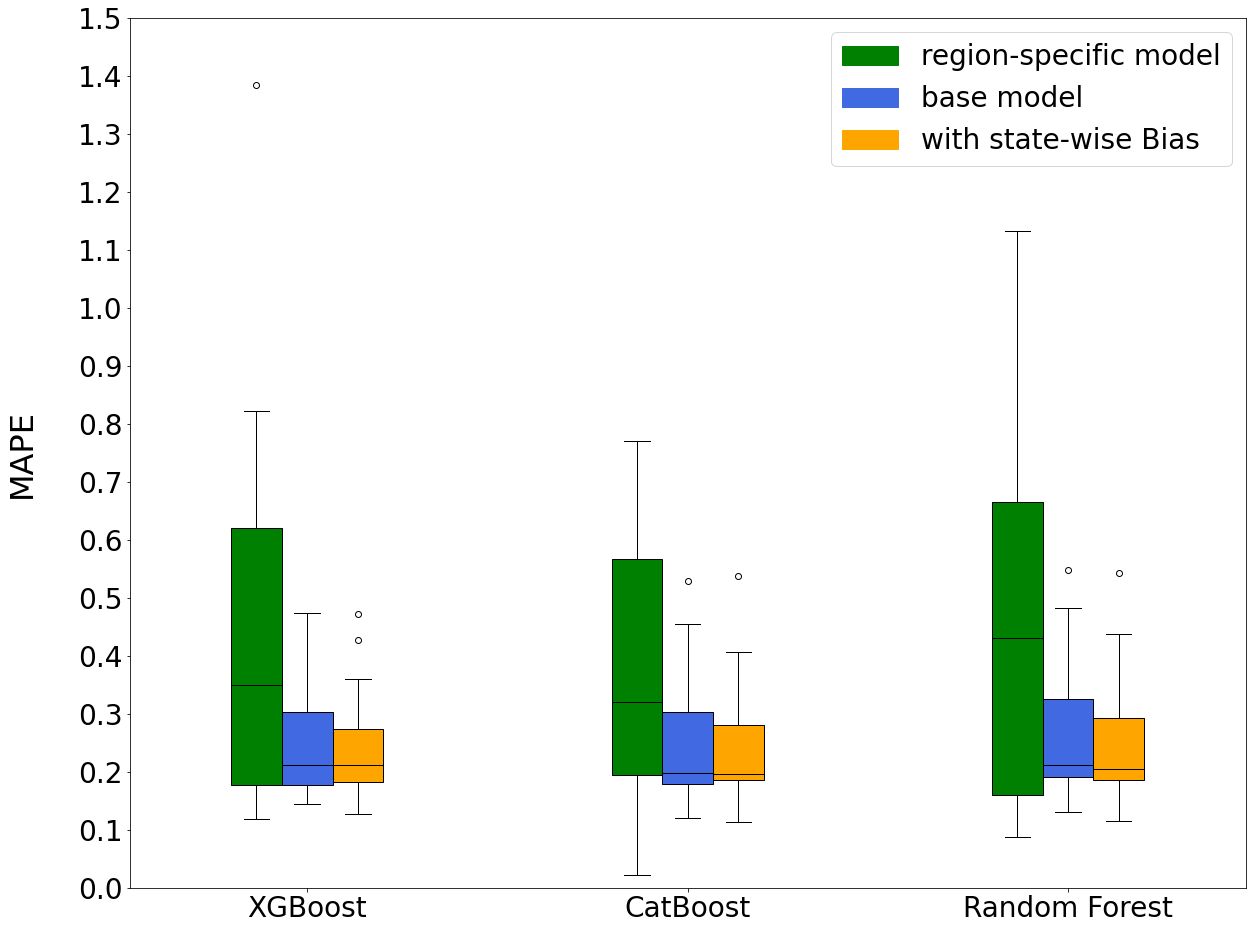}
\end{center}
\caption{Comparison between wheat yield forecast errors of different models. Each MAPE in each boxplot represents the performance of each leave-one-out test year, spanning from 2006 to 2020.}
\label{fig:boxplot}
\end{figure}
%%%%%%%%%%%%%%%%%%%%%%%%%%%%%
%%%%%%%% End Figure %%%%%%%%%
%%%%%%%%%%%%%%%%%%%%%%%%%%%%%
Besides baseline models, we also compare our approach with region-specific models, an approach that has been used in several works to forecast crop yields \cite{lee2022maize}.  
Figure~\ref{fig:boxplot} shows that when a separate model was trained on each province, the MAPE has unusually high variance (green boxplot), ranging from \textbf{5\%} to \textbf{110\%} and having a median of \textbf{40\%}. This is due to limited data available for each province (75 rows), causing a highly unstable training and serious overfitting issue for this approach. Therefore, although training a region-specific model achieved excellent performance when there are many data available, this approach is not suitable for our case. On the contrary, the performance consistently improves in all baseline models when state-wise bias is introduced (orange boxplot). Although the median MAPE was not improved by a remarkable margin (from \textbf{22\%} to \textbf{21\%}), the maximum and Q3 MAPEs are significantly decreased compared to those of the baseline models. Specifically, the maximum MAPE decreases from \textbf{48\%} to \textbf{42\%} for Random Forest, \textbf{46\%} to \textbf{41\%} for CatBoost, and \textbf{47\%} to \textbf{37\%} for XGBoost. These observations indicate that 
the proposed state-wise bias is most effective in difficult cases (cases with the highest errors) while having positive yet small impacts on easier predictions.

% Several factors, such as areas and economical policies, cause wheat yields in some states to be 2 to 3 times higher than some others. In these scenarios, RMSE is biased toward regions with higher yields, putting less weight on smaller regions. Compared to RMSE, MAPE more realistically reflects the model performance, especially when there are some large yield gaps between provinces and between years. Therefore, we use MAPE as the primary metric to compare the performance of different approaches, as shown in Figure~\ref{fig:boxplot}. 

%%%%%%%%%%%%%%%%%%%%%%%%%%%%%%%%%%%%%%%%%%%%%%%%%%%%%%%%%%%%%%%%%%
%%%%%%%%%% SECTION 4: CONCLUSION & FUTURE WORK %%%%%%%%%%%%%%%%%%%
%%%%%%%%%%%%%%%%%%%%%%%%%%%%%%%%%%%%%%%%%%%%%%%%%%%%%%%%%%%%%%%%%%

\section{Conclusion and Future Work}
Machine Learning models are frequently biased toward average yield in the dataset, resulting in higher errors for provinces with crop yields far from the mean, as shown in Figure~\ref{fig:scatterplot} This issue is exacerbated by the spatial heterogeneity between different provinces/states. Our simple state-wise bias approach can alleviate the margin of errors in such cases by ``debiasing" the errors computed separately for each province. This results in significant prediction error reduction, especially in provinces with high multi-year yields. Based upon these observations, we aim to further explore other more effective spatial-aware algorithms, such as unsupervised spatial clustering, that are robust to geospatial variations.

% \textcolor{red}{Can you help me complete the conclusion and future work? Thank you very much! } \\ \\
% \textcolor{red}{Some points that I think maybe helpful:} \\
% \textcolor{red}{1. Investigate the exact reason whether machine learning models often overfit to mean yield in other dataset and why.} \\
% \textcolor{red}{2. Based upon the knowledge from the investigation to develop more effective techniques that further reduce the errors for extreme yields.}

% \subsection{Footnotes}

% Indicate footnotes with a number\footnote{Sample of the first footnote} in the
% text. Place the footnotes at the bottom of the page on which they appear.
% Precede the footnote with a horizontal rule of 2~inches
% (12~picas).\footnote{Sample of the second footnote}

\subsubsection*{Acknowledgments}
The authors acknowledge USAID grant 720BHA21IO00261 for funding this work, as well as the efforts of our partners in FAO.

\bibliography{iclr2023_conference}
\bibliographystyle{iclr2023_conference}

\end{document}

%% file: math_commands.tex
\usepackage{amsmath,amsfonts,bm}

\def\eqref#1{equation~\ref{#1}}
\def\1{\bm{1}}

\DeclareMathAlphabet{\mathsfit}{\encodingdefault}{\sfdefault}{m}{sl}
\SetMathAlphabet{\mathsfit}{bold}{\encodingdefault}{\sfdefault}{bx}{n}

%% file: iclr2023_conference.bbl
\begin{thebibliography}{7}
\providecommand{\natexlab}[1]{#1}
\providecommand{\url}[1]{\texttt{#1}}
\expandafter\ifx\csname urlstyle\endcsname\relax
  \providecommand{\doi}[1]{doi: #1}\else
  \providecommand{\doi}{doi: \begingroup \urlstyle{rm}\Url}\fi

\bibitem[Basso \& Liu(2019)Basso and Liu]{basso2019seasonal}
Bruno Basso and Lin Liu.
\newblock Seasonal crop yield forecast: Methods, applications, and accuracies.
\newblock \emph{advances in agronomy}, 154:\penalty0 201--255, 2019.

\bibitem[Becker-Reshef et~al.(2022)Becker-Reshef, Bandaru, Barker, Coutu,
  Deines, Doorn, Eilerts, Franch, Galvez, Hosseini, et~al.]{becker2022nasa}
Inbal Becker-Reshef, Varaprasad Bandaru, Brian Barker, Sylvain Coutu, Jillian~M
  Deines, Bradley Doorn, Gary Eilerts, Belen Franch, Antonio~Sanchez Galvez,
  Mehdi Hosseini, et~al.
\newblock The nasa harvest harvest program on agriculture agriculture and food
  security food security harvest agriculture.
\newblock In \emph{Remote Sensing of Agriculture and Land Cover/Land Use
  Changes in South and Southeast Asian Countries}, pp.\  53--80. Springer,
  2022.

\bibitem[Dinh \& Aires(2022)Dinh and Aires]{gmd-15-3519-2022}
T.~L.~A. Dinh and F.~Aires.
\newblock Nested leave-two-out cross-validation for the optimal crop yield
  model selection.
\newblock \emph{Geoscientific Model Development}, 15\penalty0 (9):\penalty0
  3519--3535, 2022.
\newblock \doi{10.5194/gmd-15-3519-2022}.
\newblock URL \url{https://gmd.copernicus.org/articles/15/3519/2022/}.

\bibitem[FAO(2022)]{FAOSTAT2022}
FAO.
\newblock Faostat: Fao statistical databases 2022, 2022.
\newblock URL \url{https://www.fao.org/faostat/en/#home}.

\bibitem[Lee et~al.(2022)Lee, Davenport, Shukla, Husak, Funk, Harrison,
  McNally, Rowland, Budde, and Verdin]{lee2022maize}
Donghoon Lee, Frank Davenport, Shraddhanand Shukla, Greg Husak, Chris Funk,
  Laura Harrison, Amy McNally, James Rowland, Michael Budde, and James Verdin.
\newblock Maize yield forecasts for sub-saharan africa using earth observation
  data and machine learning.
\newblock \emph{Global Food Security}, 33:\penalty0 100643, 2022.

\bibitem[Meroni et~al.(2021)Meroni, Waldner, Seguini, Kerdiles, and
  Rembold]{MERONI2021108555}
Michele Meroni, François Waldner, Lorenzo Seguini, Hervé Kerdiles, and Felix
  Rembold.
\newblock Yield forecasting with machine learning and small data: What gains
  for grains?
\newblock \emph{Agricultural and Forest Meteorology}, 308-309:\penalty0 108555,
  2021.
\newblock ISSN 0168-1923.
\newblock \doi{https://doi.org/10.1016/j.agrformet.2021.108555}.
\newblock URL
  \url{https://www.sciencedirect.com/science/article/pii/S0168192321002392}.

\bibitem[Sahajpal et~al.(2020)Sahajpal, Fontana, Lafluf, Leale, Puricelli,
  O’Neill, Hosseini, Varela, and Becker-Reshef]{sahajpal-argentina}
Ritvik Sahajpal, Lucas Fontana, Pedro Lafluf, Guillermo Leale, Estefania
  Puricelli, Dan O’Neill, Mehdi Hosseini, Mauricio Varela, and Inbal
  Becker-Reshef.
\newblock Using machine-learning models for field-scale crop yield and
  condition modeling in argentina.
\newblock 2020.
\newblock \doi{10.31223/x52595}.

\end{thebibliography}
